\documentclass{article}

\usepackage[final]{nips_2016}

\usepackage[utf8]{inputenc} 
\usepackage[T1]{fontenc}    
\usepackage{hyperref}       
\usepackage{url}            
\usepackage{booktabs}       
\usepackage{amsfonts}       
\usepackage{nicefrac}       
\usepackage{microtype}      
\usepackage{graphicx}
\usepackage{natbib}
\usepackage{wrapfig}
\usepackage{caption}
\usepackage{subcaption}

\title{Model-driven Simulations for Deep Convolutional Neural Networks}

\author{
  V S R Veeravasarapu$^1$, Constantin Rothkopf$^2$, Visvanathan Ramesh$^1$\\ 
 $^1$Center for Cognition and Computing, Goethe University, Frankfurt.\\
 $^2$Cognitive Science Center,  Technical University, Darmstadt.
}

\begin{document}

\maketitle

\begin{abstract}

The use of simulated virtual environments to train deep convolutional neural networks (CNN) is a currently active practice to reduce the (real)data-hungriness of the deep CNN models, especially in  application domains in which large scale real data and/or groundtruth acquisition is difficult or laborious. Recent approaches have attempted to harness the capabilities of existing video games, animated movies to provide training data with high precision groundtruth. However, a stumbling block is in how one can certify generalization of the learned models and their usefulness in real world data sets. This opens up fundamental questions such as: \textit{What is the role of photorealism of graphics simulations in training CNN models?  Are the trained models valid in reality?  What are possible ways to reduce the performance bias?} In this work, we begin to address theses issues systematically in the context of urban semantic understanding with CNNs. Towards this end, we (a) propose a simple probabilistic urban scene model, (b) develop a parametric rendering tool to synthesize the data with groundtruth, followed by (c) a systematic exploration of the impact of level-of-realism on the generality of the trained CNN model to real world; and domain adaptation concepts to minimize the performance bias.  
\end{abstract}

\section{Introduction}

The use of virtual environments to train CNN-based vision models is starting to be active practice to reduce the data-hungriness of the deep models, especially in application domains where large scale real world data acquisition and groundtruth collection is laborious or difficult. A recent example  harnesses the power of gaming engines in a automotive racing context and animated movies to provide large scale datasets along with high precision groundtruth (\cite{chen2015deepdriving, haltakov2013framework}).  The simulated data from these gaming engines are limited to specific contexts and the real-time gaming engine requirement necessitates the use of approximations in graphics rendering.  Model-based graphics simulations are increasingly integrated into machine learning \cite{vazquez2014virtual, haltakov2013framework} or probabilistic programming platforms \cite{kulkarnipicture} that allow numerical simulations to be tightly integrated into learning and inference. We propose to 
use stochastic generative models (in 3D), physics-inspired rendering methods and exploit the availability of online repositories of 3D CAD object shapes and textures (for instance, Google's sketchup 3D warehouse) to synthesize diverse virtual environments.

Despite advances in computer graphics tools, there are still several issues that have to be kept in mind while using it for training purposes.  Exact sampling and rendering processes from physics inspired models are still mostly infeasible and computationally intractable. Hence, approximations using Monte-carlo methods to solve intractable integrals involved in the graphics rendering process. The accuracy of the rendered outputs would depend on number of samples used to solve the integrals. We refer the accuracy of rendering as \textbf{\textit{level-of-fidelity}}. Stochastic modeling for a given application domain (especially in 3D) is about describing the probability distribution over 3D geometric structures and appearances of scenes and objects with various degrees of abstractions. Since all models are simplifications of reality there is always a trade-off as to what level of detail is included in the model. If too little detail is included in the model one runs the risk of getting very unrealistic images. If too much detail is included in the model the model may become overly complicated and becomes intractable. We refer this as \textbf{\textit{level-of-detail}}. In general, the photo-realism of the rendered outputs and its statistics depend on level-of-detail that is encoded in the stochastic domain models (and their closeness to real-world) and level-of-fidelity of rendering engine.

Stochastic point processes \cite{stoyan1994fractals} have been used to model geo-spatial distributions \cite{gatrell1996spatial} of aerial image analysis. In this work, we use a special class of point processes, names Marked Poisson Processes (MPP) along with current graphics simulation tools to model the image generation process and to simplify the large scale database preparation for training the CNNs for urban scene understanding. However, one major stumbling block is in certification of the designs and tuned/learnt parameters of the scene understanding system to work in real world datasets, given the fact that level-of-detail in the domain models and level-of-fidelity in the renderings are limited.  This opens up fundamental questions such as 
\textit{What is the role of photo-realism of graphics simulations in training CNN models ?  
Are these models valid in reality?  
How can we correct for the bias introduced in training due to discrepancy between statistics of  simulated data and real world data? }

We begin to systematically explore these aspects in the context of semantic segmentation of urban street scene images. The main contributions of our work are (i) A simple and effective probabilistic scene model for 3D urban street scenes (ii) A rendering platform to generate data with required groundtruth (iii) Systematic experimentation to analyze the importance of photo-realism for generalization of the network trained, and exploration of ways to correct the bias in the models learnt in virtual worlds. In the Section \ref{sec_review}, we review past work that use graphics simulations in the vision system design process, and related literature in urban scene segmentation. Section \ref{sec_pgm} discusses our stochastic domain models and the rendering platform. Experiments to validate our models for training DCNs are discussed in Section \ref{sec_exps}. Section \ref{sec_discuss} concludes with a discussion on our analysis and some of future directions.

\section{Background}\label{sec_review}

\textbf{Graphics for Vision}:- The question of whether graphics simulations are in fact realistic enough for computer vision has been addressed in past work. For instance, \cite{vaudrey2008differences,vazquez2014virtual,DBLP:journals/pami/XuRVL14} used basic rendering algorithms and scene parameters and concluded that simulations alone are not useful for tuning the respective vision systems, where as the work \cite{meister2011real} used carefully designed indoor scene models (parameters) and advanced rendering algorithms to synthesize very realistic sensory data and concluded that graphics can be used. The work \cite{butler2012naturalistic} showed that motion models and local spatial statistics, crucial for optic flow estimation, match with reality. Hence, they argued that the artificial animated (Sintel) data could be used to design and tune the flow estimators even though the data is not photo-realistic. Recently, \cite{fischer2015flownet} used a combination of real-data and synthetic objects to train a DCN and provide empirical evidence of the usefulness of such training for real world settings. The main focus of these works is in demonstration of the utility or lack of utility of simulation for vision systems design and the emphasis is on evaluation of the system as a black-box in an application context. Still, the degree of effectiveness of graphics rendered data for vision system design is an open question.  A discussion \footnote{www.quora.com/How-useful-are-massive-virtual-game-environments-for-training-AI} on a social network platform clearly conveys the confusion that the vision community have towards using graphics for learning. Hence, we believe that a systematic exploration of this space should be done in the context of current state-of-the-art graphics tools.

\textbf{Graphics for Urban Scene Understanding}:- Racing video game engines, (V-drift, Torcs and H-life etc.), have been used to produce synthetic annotated data for vision for automotive applications. The work \cite{haltakov2013framework} modified V-drift to synthesize the ground-truth video data for the task of semantic segmentation. The work evaluated different input data structures (RGB, Depth, Flow etc) achieve the better performance for driver assistance systems. The work \cite{vazquez2014virtual} use H-life game engine to generate synthetic data to train several pedestrian detectors. They also demonstrated domain adaptation methods by exploring several ways of combining a few real world pedestrian samples to many synthetic samples from H-life environments. Recent work of \cite{chen2015deepdriving} trained a CNN using 12 hours of human driving in Torcs video game and showed that the CNN models can work well to drive a car in a very diverse set of virtual environments. However, these game engines were developed with real time rendering for gaming purposes. In general, most of the effects such as fog scenes, were mathematically simplified \cite{pharr2004physically}. Also, the diversity of environments in the racing engines is also quite limited. As a next step, we use stochastic Poisson processes with 3D CAD designs and advanced rendering tools to create diversified and realistic large scale data along with ground-truth. 

\textbf{CNNs for Urban Scene Understanding}:- With the successful adaptation of deep networks from classification to pixel-level prediction tasks, a wide variety of CNN architectures for semantic segmentation have been proposed. One class of architectures focus on harnessing the power of model driven designs (graphical models of context) into CNNs, for instance DeepLab \cite{chen2014semantic} and CRF-RNN \cite{zheng2015conditional}. These approaches usually constrain local level predictions from CNNs (unary potentials) with a graphical model (such as conditional random fields), in which priors and constraints on global relations (n-ary potentials) are incorporated. These models are then trained to maximize the likelihood of correct classification given the features. Several attempts to collect large scale real world data sets for use in benchmarks have been reported \cite{cordts2016cityscapes}, however, these data sets are collected by using a particular sensor and viewpoint settings(located in a car). Transferability of the models trained on these datasets to new data is often questionable if the experimental design for data collection is not systematic and the data reflects a wide range of acquisition scenarios.  Graphics simulations can allow for changing sensor properties, view points, etc. (See for instance, \cite{hattori2015learning}). 

\textbf{Bias in Performance via training using Simulated data}:- 
The significance of the impact of level-of-fidelity of simulations on the bias in performance of the trained model, when applied on real-world data, depends on (a) level of details encoded in the domain models , (b) level of fidelity of rendering engine, and (c) invariance of the modules/system to the parameters of scene photometry and graphics engine. 
A more elaborated literature survey on these research areas can be found in \cite{veeravasarapu2015simulations}. 
This perspective motivates us to develop a simulation platform which facilitates probabilistic domain modeling and parametric rendering along with the required groundtruth. The next section describes the details of our domain models and simulations platform.

\begin{figure}
\centering
\includegraphics[width=14cm, height=4cm]{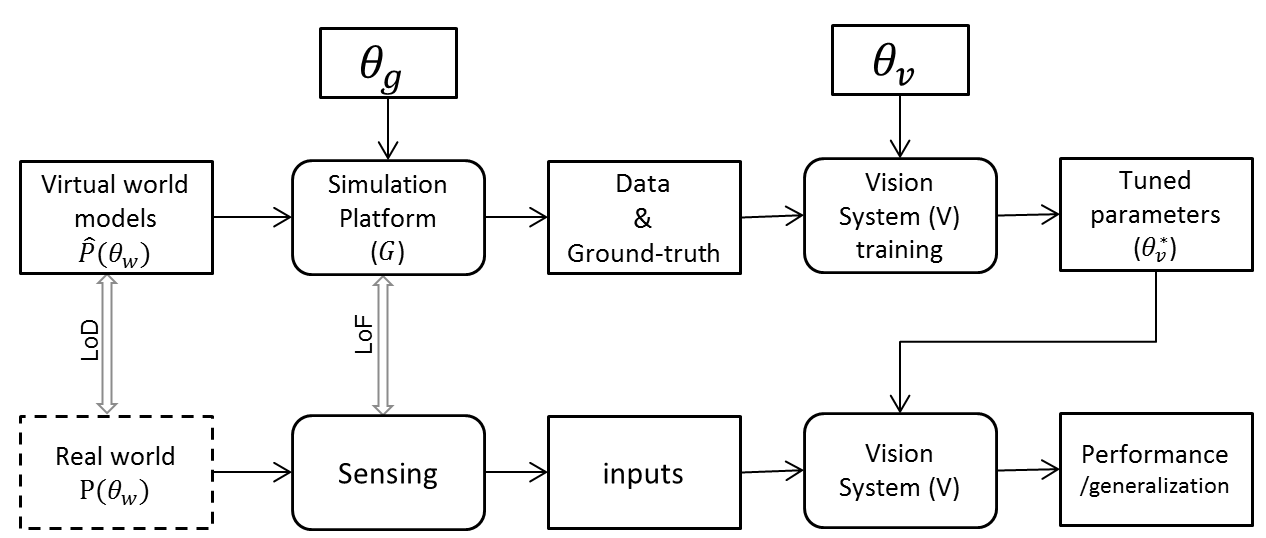}
\caption{Distribution propagation from virtual domains to real world performance}
\label{fig_distr}
\end{figure}

\section{Application Domain Modeling} \label{sec_pgm}
\begin{figure}
    \centering
    \begin{subfigure}[b]{0.48\textwidth}
        \centering
        \includegraphics[width=7cm, height=4cm]{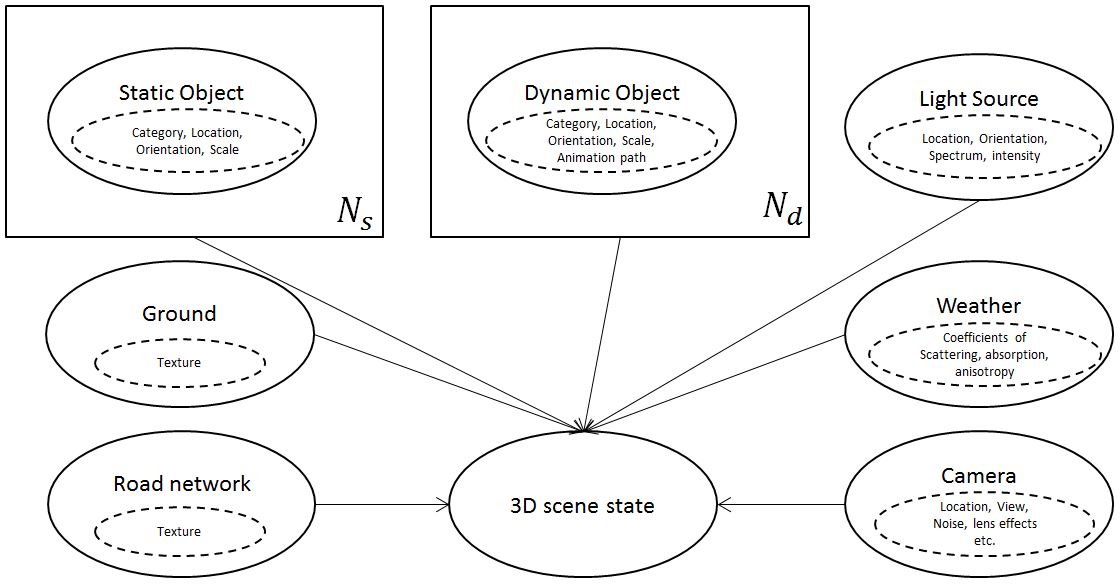}
        \caption{Domain model: rectangle boxes represent MPP, $N_s$ and $N_d$ denote the number of static and dynamic objects respectively in the virtual world} 
        \label{fig_pgm}
    \end{subfigure}
    \hfill
    \begin{subfigure}[b]{0.48\textwidth}
        \centering
        \includegraphics[width=7cm, height=5cm]{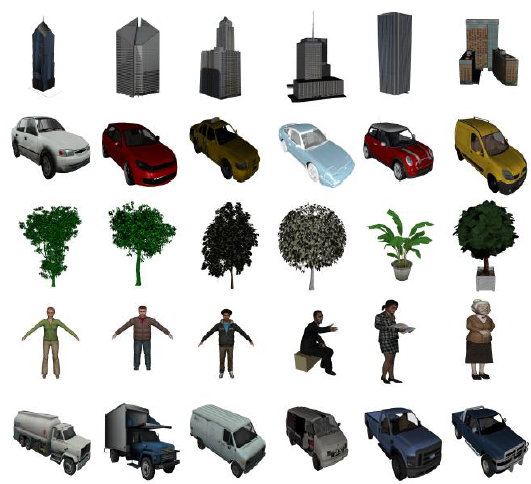}
        \caption{3D CADs as marks library  }
        \label{fig_marks}
    \end{subfigure}
    \hfill
    \begin{subfigure}[b]{1.0\textwidth}
        \centering
        \includegraphics[width=0.245\textwidth, height=2cm]{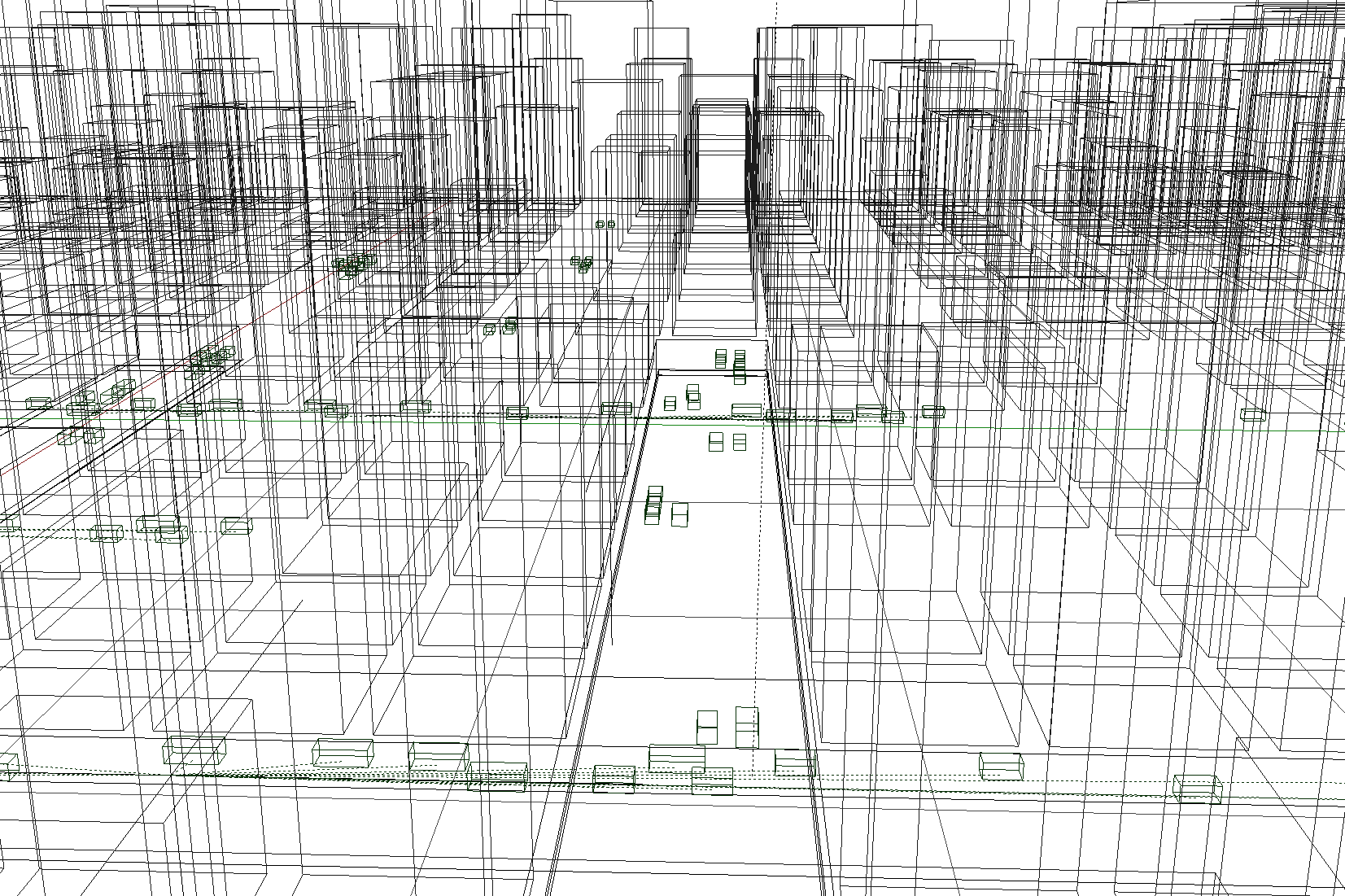}
        \includegraphics[width=0.245\textwidth, height=2cm]{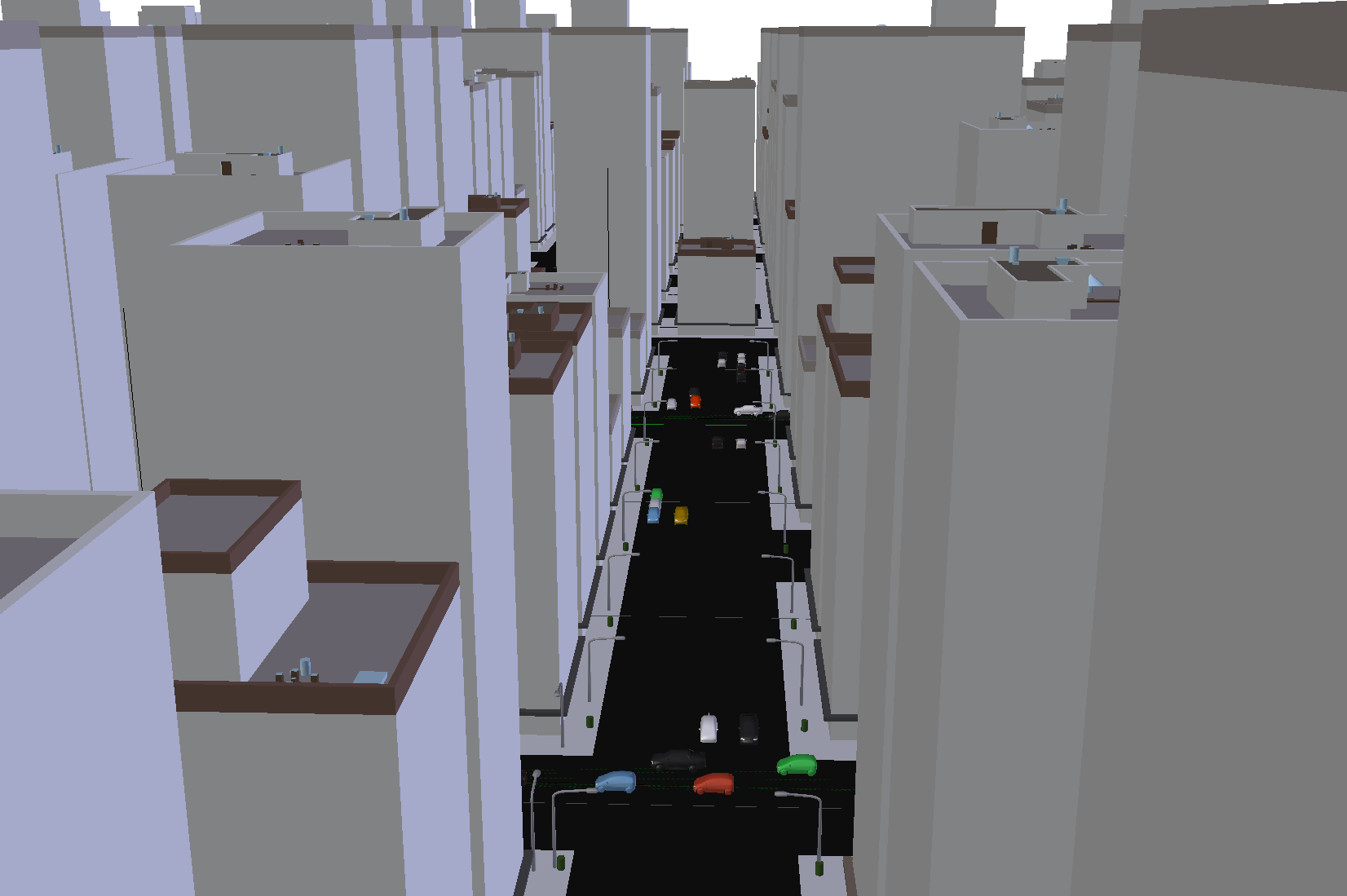}
        \includegraphics[width=0.245\textwidth, height=2cm]{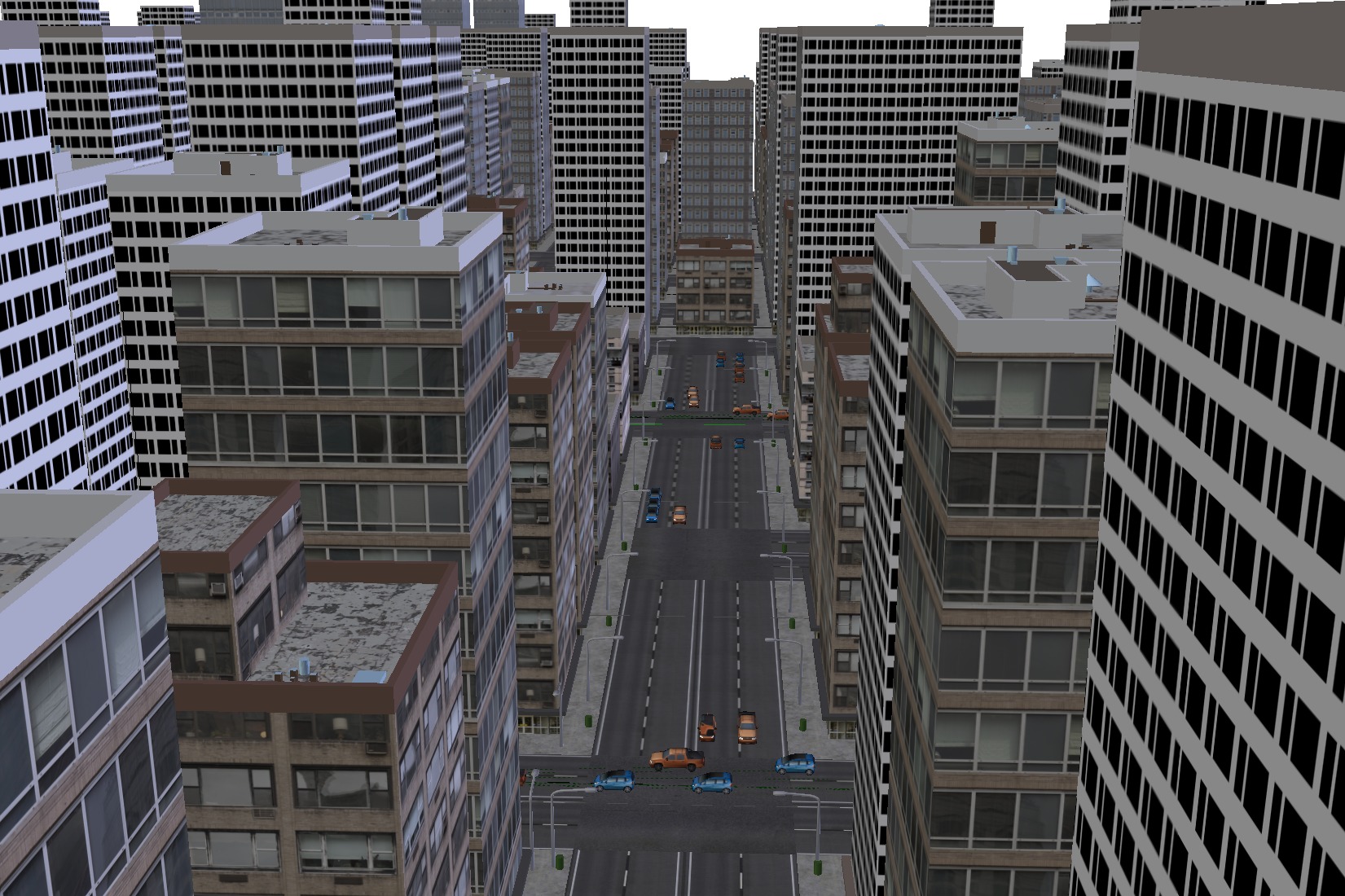}
        \includegraphics[width=0.245\textwidth, height=2cm]{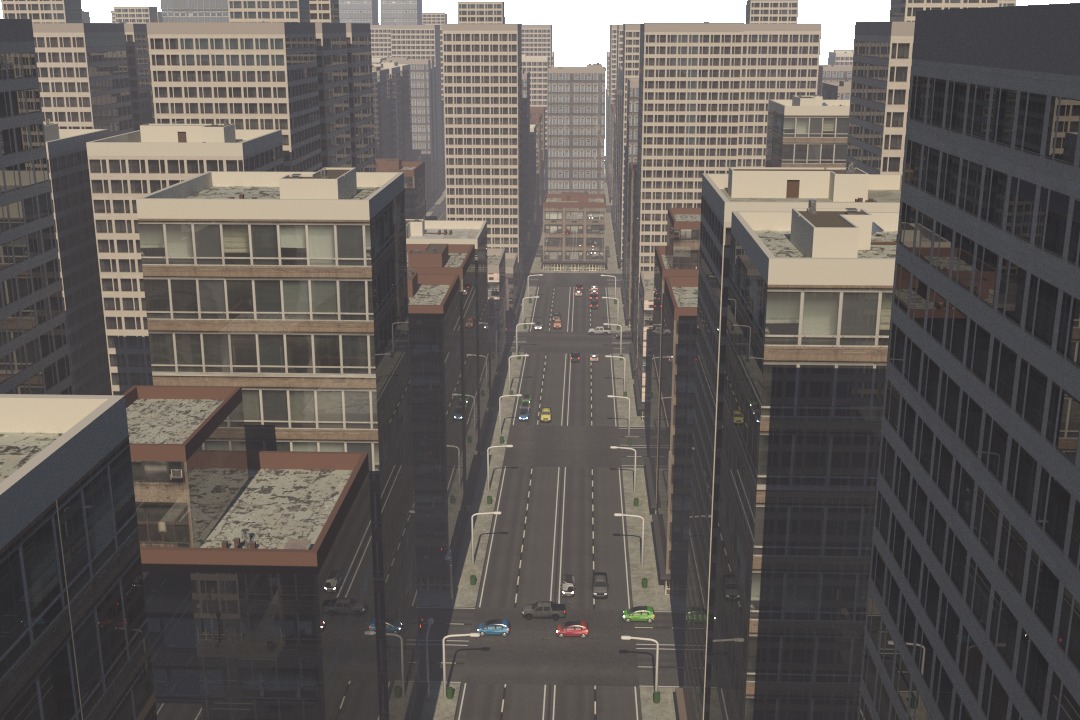}
        \caption{Bird-eye view of a urban scene at various stages of synthesis process}
        \label{fig_mpp}
    \end{subfigure}
    
    \begin{subfigure}[b]{1.0\textwidth}
        \centering
        \includegraphics[width=0.245\textwidth,height=2cm]{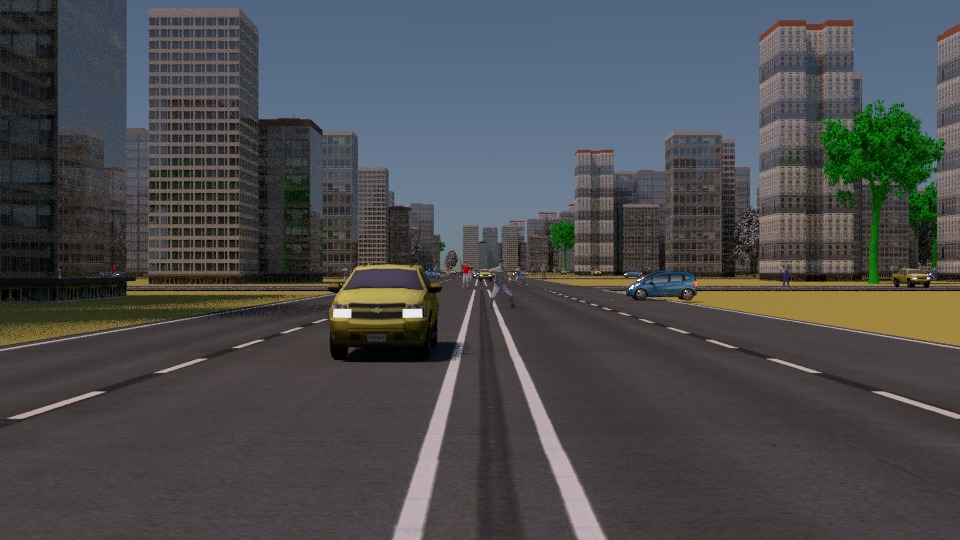}
        \includegraphics[width=0.245\textwidth,height=2cm]{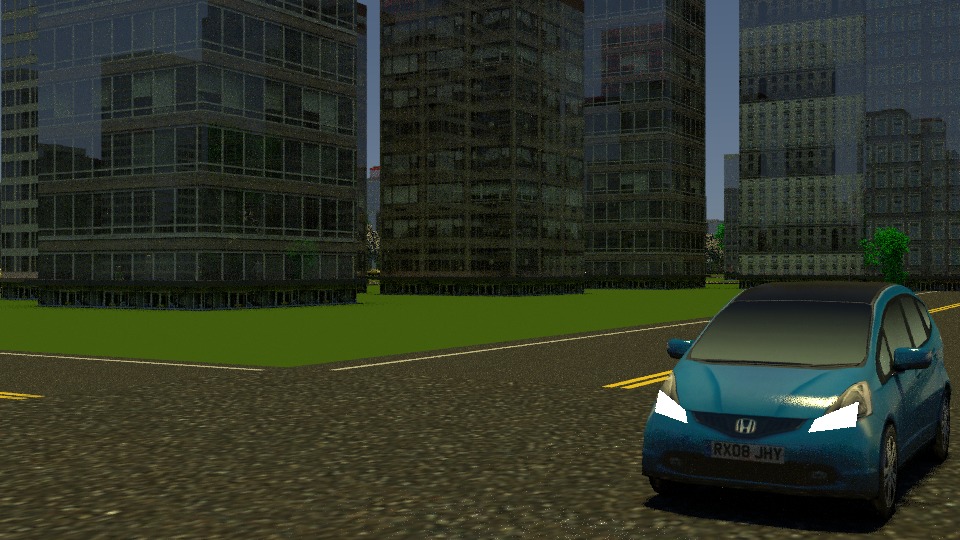}
        \includegraphics[width=0.245\textwidth,height=2cm]{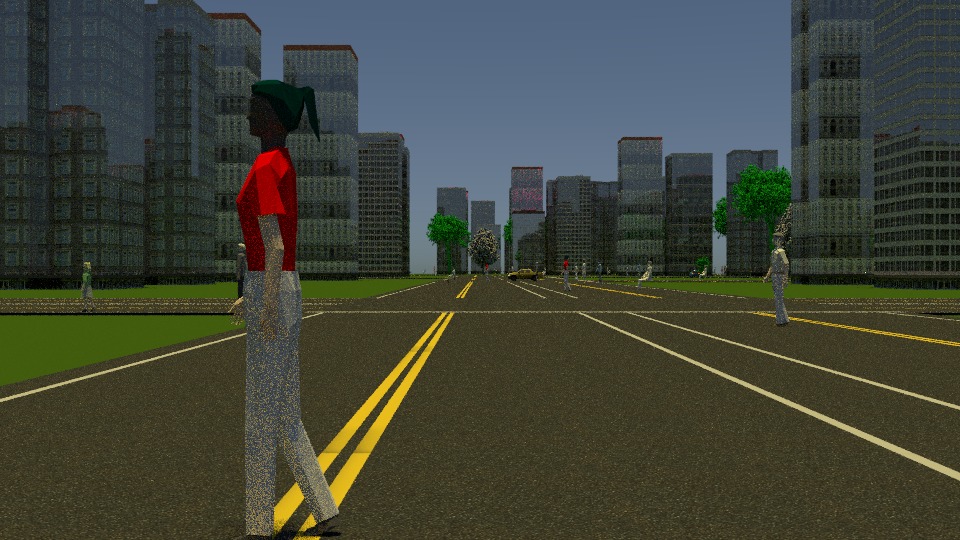}
        \includegraphics[width=0.245\textwidth,height=2cm]{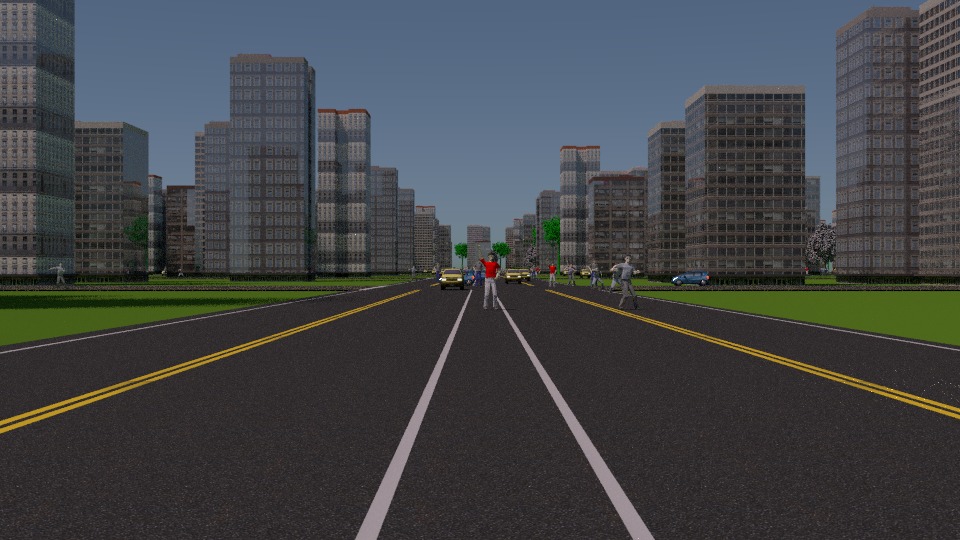}
        \caption{Geometric variations from the scene model}
        \label{fig_geo_var}
    \end{subfigure}
    
    \begin{subfigure}[b]{1.0\textwidth}
        \centering
        \includegraphics[width=0.245\textwidth,height=2cm]{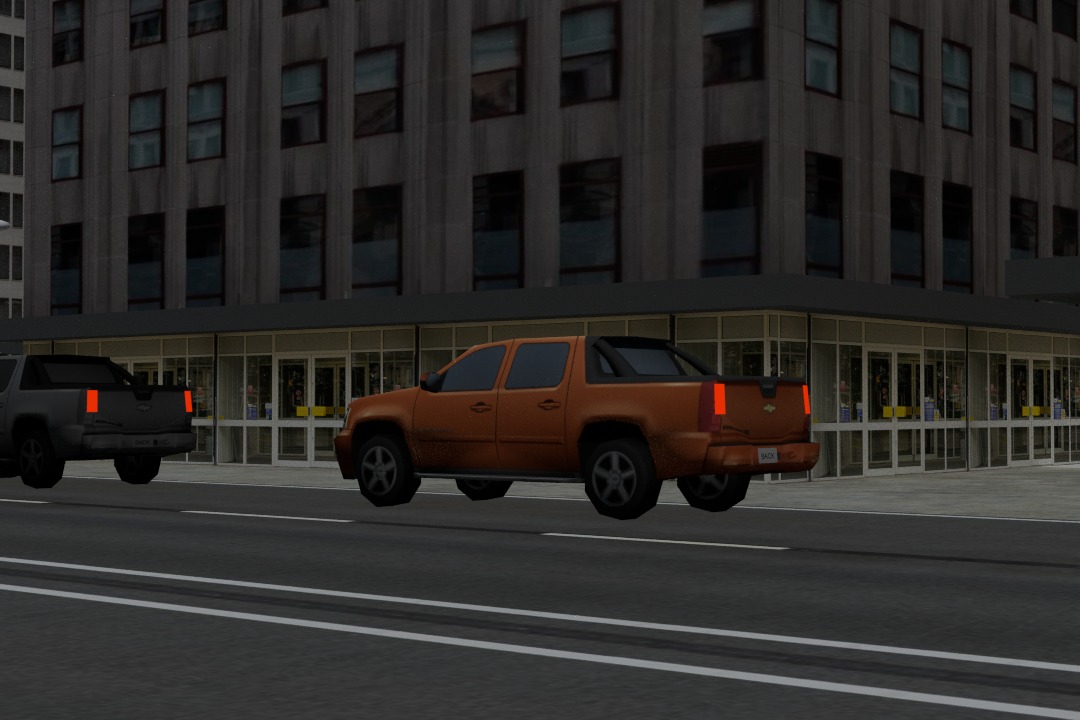}
        \includegraphics[width=0.245\textwidth,height=2cm]{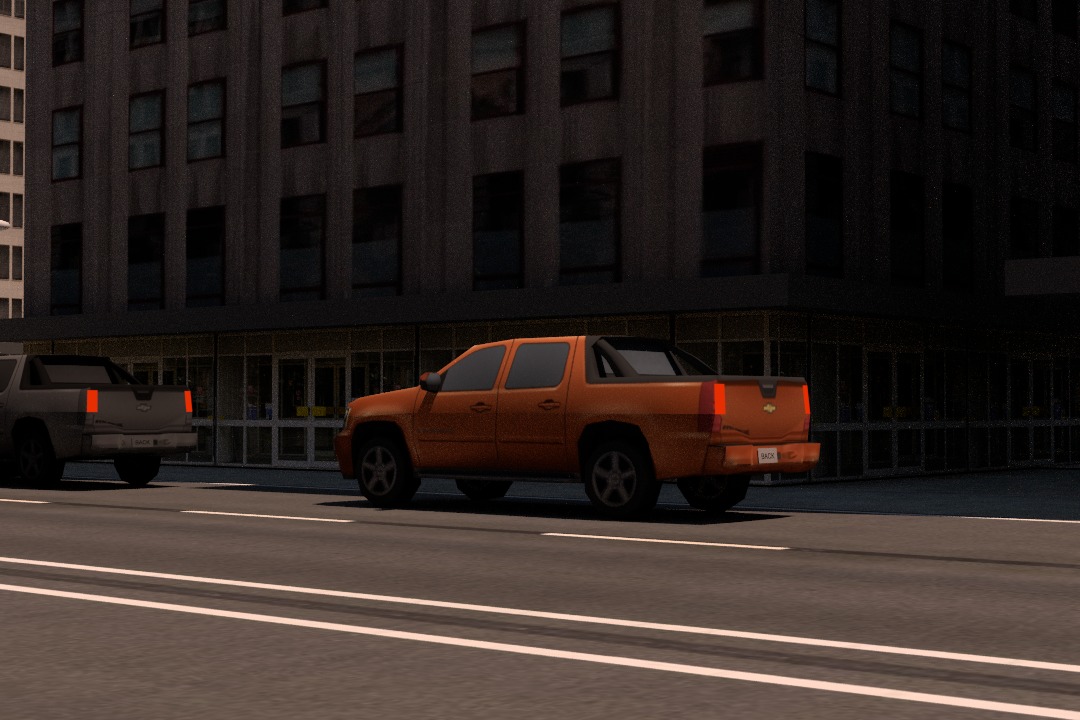}
        \includegraphics[width=0.245\textwidth,height=2cm]{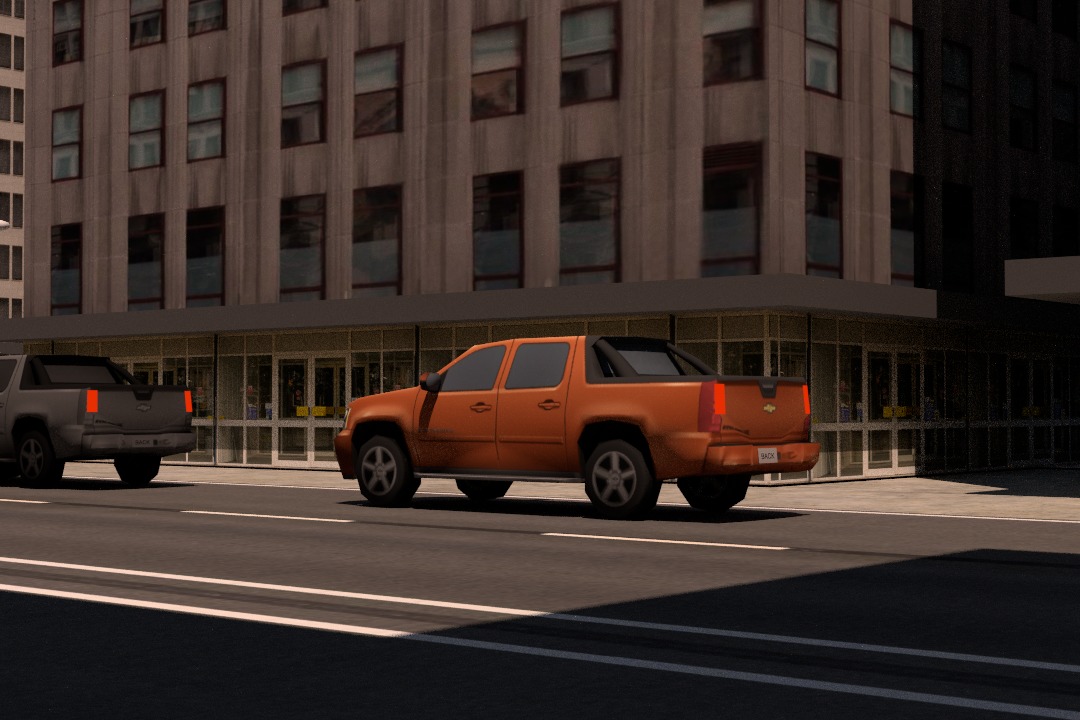}
        \includegraphics[width=0.245\textwidth,height=2cm]{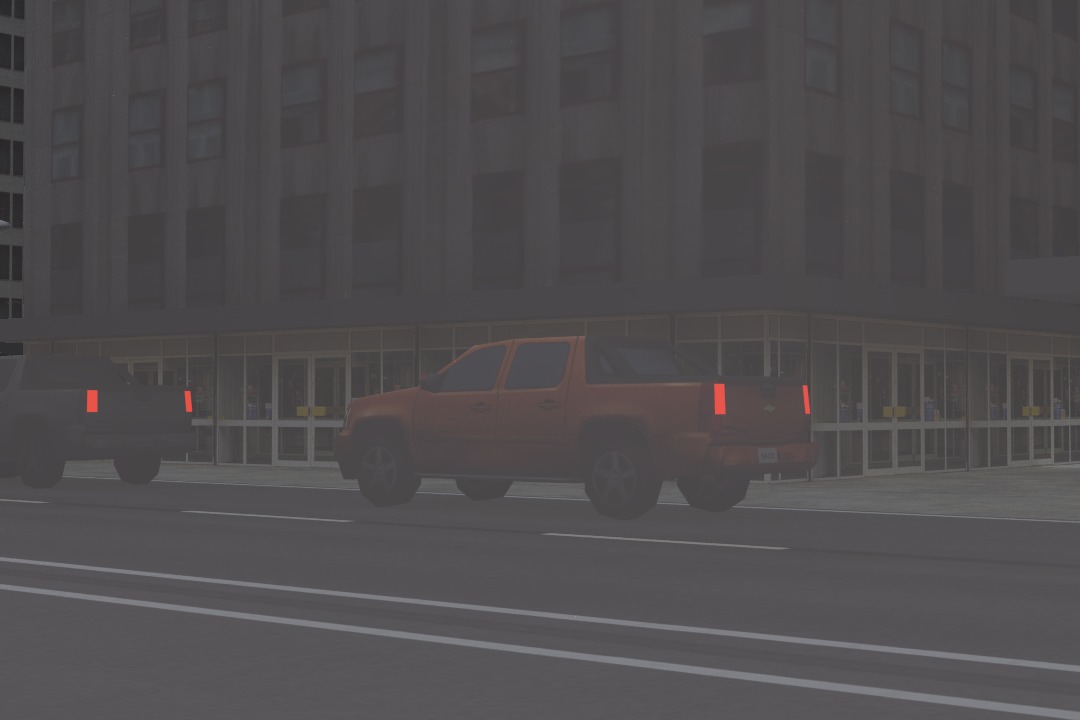}
        \caption{Photometric variations from the scene model}
        \label{fig_photo_var}
    \end{subfigure}
    \caption{Domain Modeling}
    \label{fig_dm}
\end{figure}

In this section, we model geometric, photometric and dynamical aspects of 3D urban street scenes. A graphical model which represents the semantics of our model is shown in Fig \ref{fig_pgm}. We mainly focus on geometry part in this section, please see \cite{veeravasarapu2015simulations, veeravasarapu2015model} for more elaborated discussions on photometric and dynamical models.

\subsection{Geometry Modeling with Marked Poisson Processes}

We model the ground and road network in the 3D scene with predefined planes in the world coordinate system, however, the textures are loaded randomly from our collection. We use a spatial point process, called marked poisson process (MPP), to model spatial distributions of a set of objects (such as building, vehicles etc.). MPP is one of the tools that popularly used in geo-spatial analytics such as urban, forest statistics \cite{lafarge2010geometric}. We view scene entities (objects such as building, trees, vehicles etc.) as points with marks, i.e. attributes, such as category, mesh model, textures, scale, orientation etc.  

A realization of the point process $Z$ consists of a random countable set of points with their locations $\{ o_1,...,o_n \}$ in a bounded region $\textbf{O}$. MPP couples a spatial point process $Z$ with a second process defined over a \textit{mark} space $M$ such that a random mark $m_o \in M$ is associated with each point $o \in Z$. MPP can be extended as multi-object marked point process \cite{lafarge2010geometric} to consider different object types as shown in Fig \ref{fig_mpp}. For example, a MPP of static objects (buildings, trees) on ground space has elements of the form $o_i = (p_i, ( c_i, idx_i, s_i, \theta_i))$ specifying the location ($p$), object class ($c$), index of 3D mesh and texture-maps to import ($idx$), scale ($s$) and orientation ($\theta$) of a specific static object in the scene. We instantiate two parallel MPPs, one for static objects, the other for dynamic objects distributions. For the static object MPP, the bounded space (\textbf{O}) for points is the ground plane, where as, the road network mesh corresponds to the bounded space for the dynamic object MPP. Thus, a realization of MPP is a set points (locations to place objects), together with marks of object category, index to mesh and texture files to import, scale, and orientation.  For dynamic objects, a destination point is also incorporated in the MPP. Path generation algorithms could be used to animate the objects from birth (i.e. start) location to destination.

One can choose prior probability distributions ($\pi$) on the point densities ($p$) and marks ($m$) depending on the type of scenes interested (residential, industrial etc.). We use homogeneous Poisson processes and assume independence between marks for simplicity in sampling. Hence, the priors on MPP would be factored as: $\pi(o)=\pi(p)\pi(m)$. Spatial overlap between locations of objects is minimized Gibbs energies of repulsive potentials. Likewise, several constraints or priors can be added to the model by using Gibbs densities of the form $e^{-U(x)}$, where $U$ is a Gibbs energy associated with configuration space. However, this increases the complexity in sampling process. The parameters of these point processes and conditional mark processes can be learnt from the real world data using traditional machine learning algorithms. For example, density, populations of objects in a city can be learnt from open street maps \cite{haklay2008openstreetmap}. However, in this work, we use uniform distributions on most of the parameters of MPPs.  We use online repositories, such as Google's 3D warehouse, to collect a wide variety of 3D CAD object meshes and textures for object category. 


\subsection{Photometry}

\textit{Lights and BRDFs}:- Several light models, materials (lambertian and specular), weather scattering models is already available in open source graphics rendering platforms such as {\it Blender}, which is base of our simulation platform. A light model (for sun light) which is parametrized by intrinsic variables (such as color spectrum, intensity) and extrinsic variables (such as location and orientation) is used for the simulations in this work. 
Physics inspired BRDF (Bi-directional reflectance distribution functions) models for lambertian, specular and glassy surfaces are available with Blender. We also use binary image maps (in addition to texture maps) to activate corresponding shader (lambertian or specular surface reflections). 
\textit{Weather models}:- Light that travels through weather or any other participating medium, undergoes three kinds of phenomena: absorption, in and out scattering, and emission. To simulate these effects, we use the shaders of volume scattering, absorption, which are parametrized by the coefficients of scattering, particle density and anisotropy.  Uniform distributions are used on most of the photometric variables. 
\textit{Sensor models}:- Standard OpenGL pinhole camera can be modified to meet any camera model settings. Some of the noise and lens effects (chromatic aberration, vignetting, lens distortion, sensor noise) can be modelled using post-processing shaders on the framebuffer. 
Automatic random animation and path generation for dynamic objects can be done with algorithms such as A-star, artificial potential fields etc.

\begin{figure}[h]
\centering
\includegraphics[width=3.92cm,height=2.3cm]{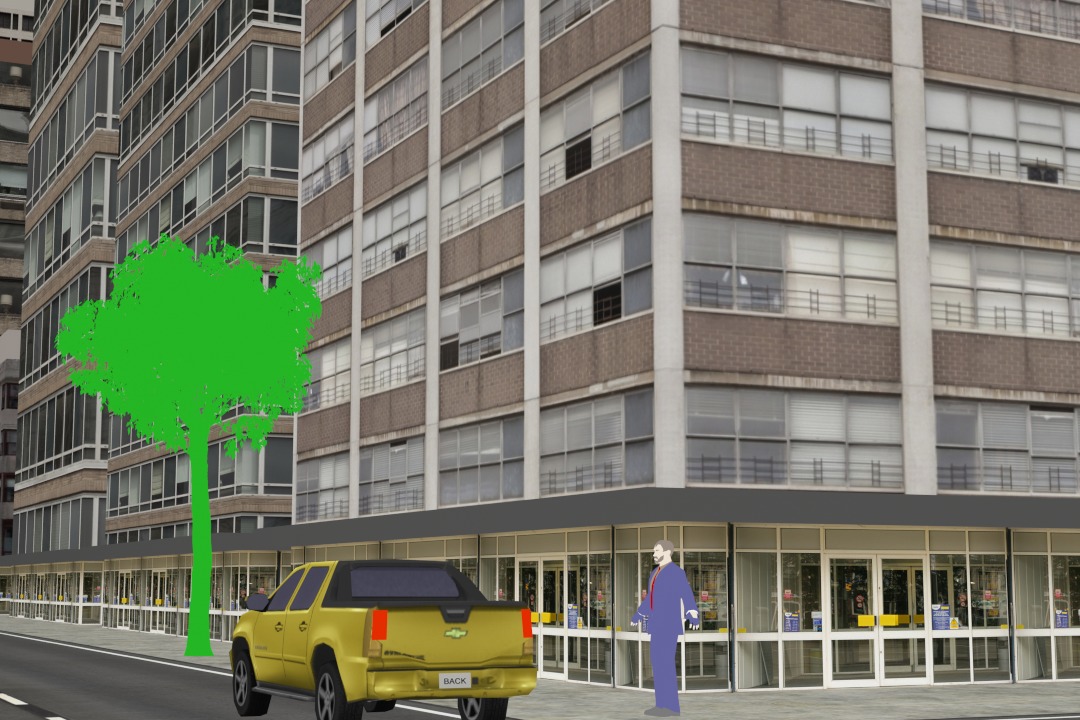}
\includegraphics[width=3.92cm,height=2.3cm]{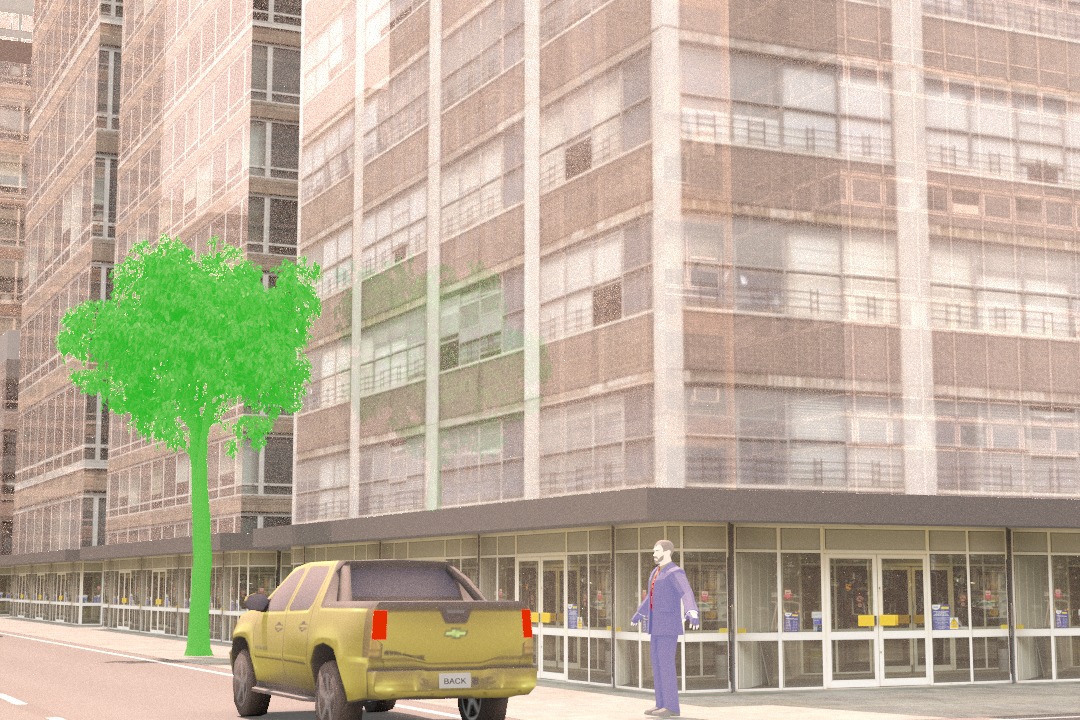}
\includegraphics[width=3.92cm,height=2.3cm]{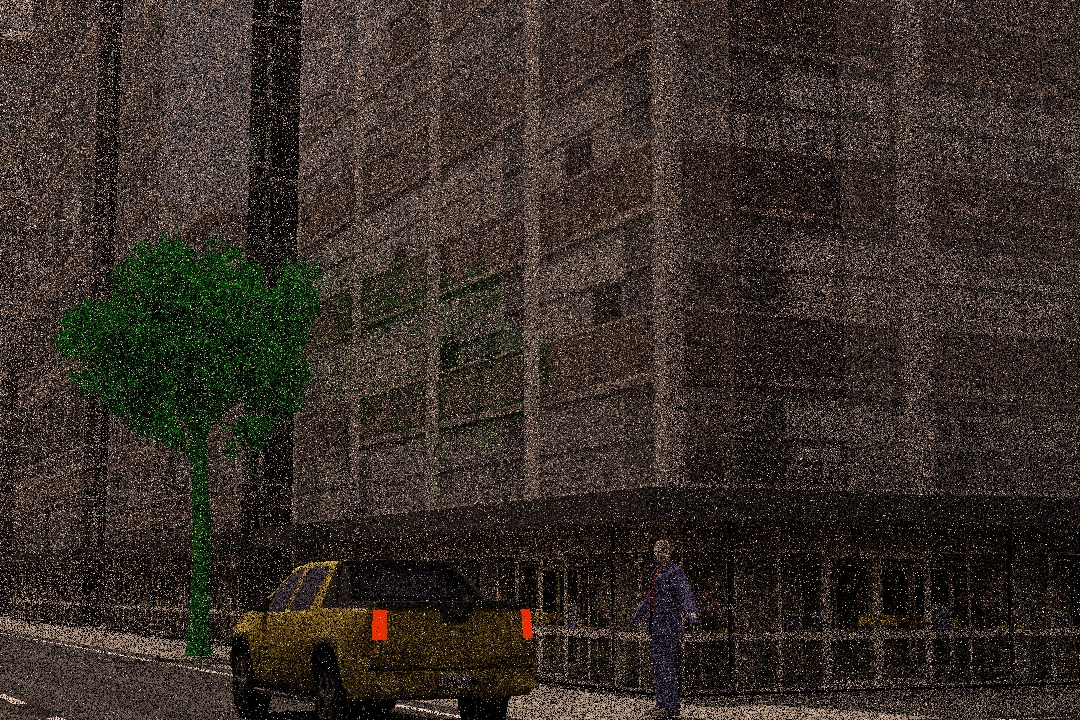} \\
\includegraphics[width=3.92cm,height=2.3cm]{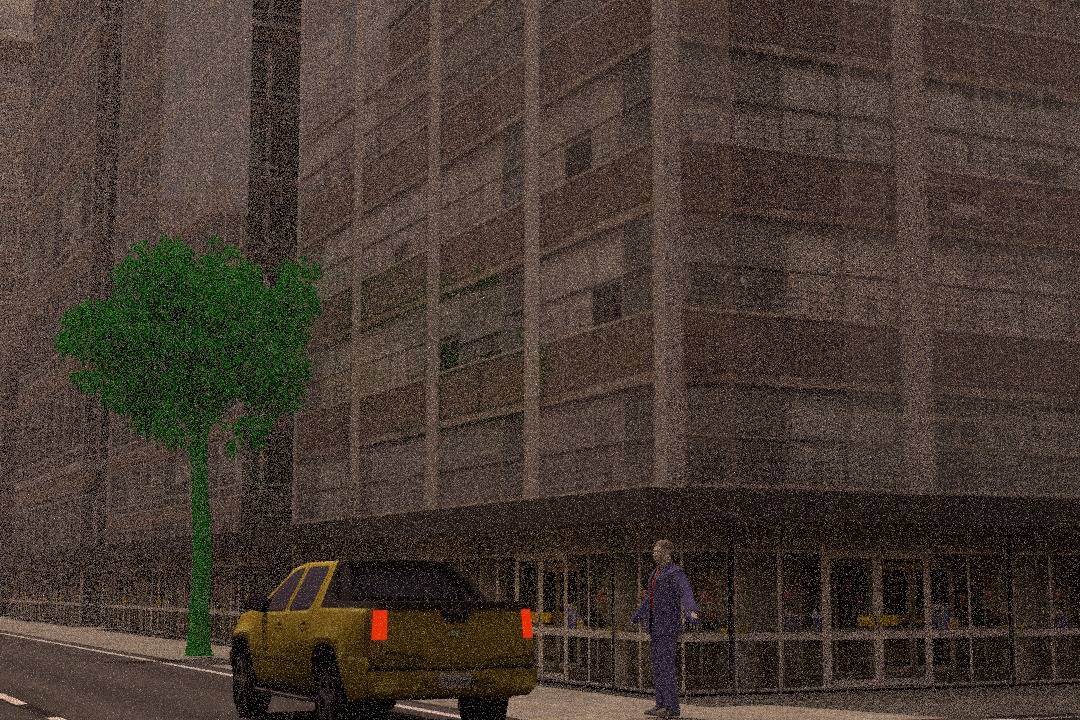}
\includegraphics[width=3.92cm,height=2.3cm]{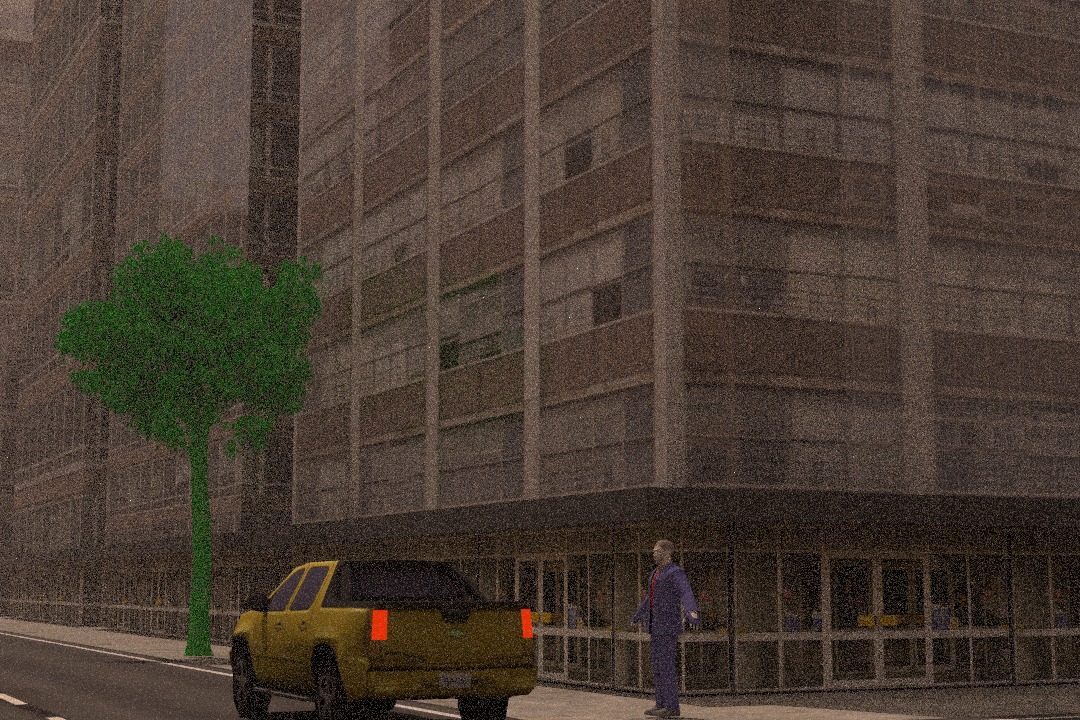}
\includegraphics[width=3.92cm,height=2.3cm]{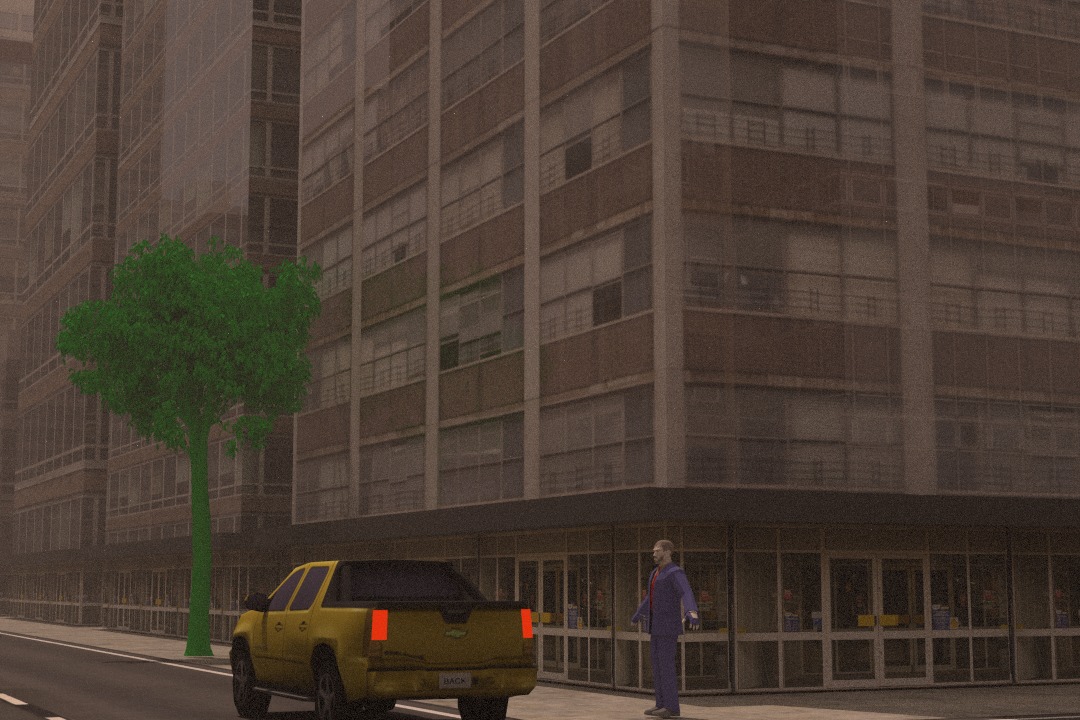} \\
\includegraphics[width=3.92cm,height=2.3cm]{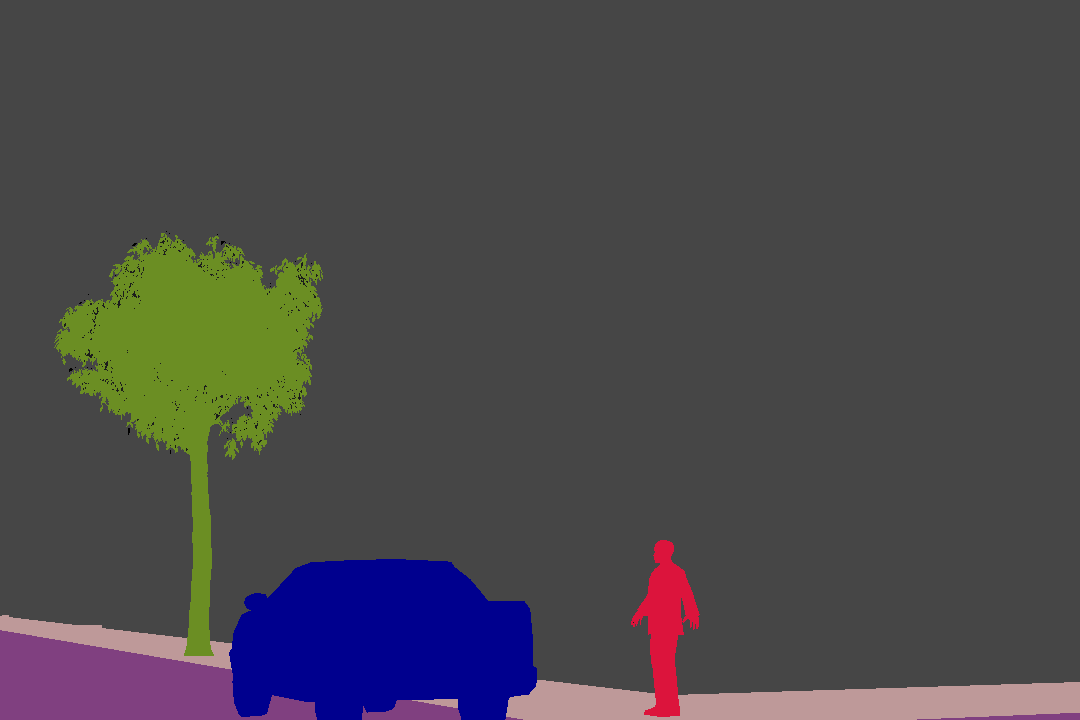}
\includegraphics[width=3.92cm,height=2.3cm]{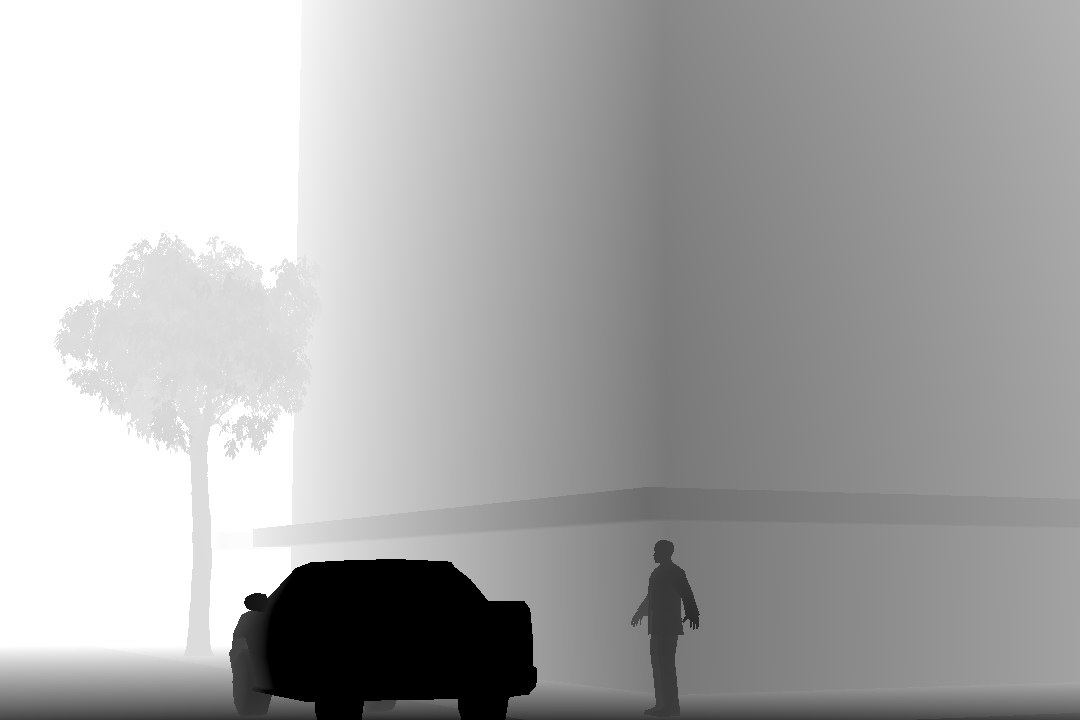}
\includegraphics[width=3.92cm,height=2.3cm]{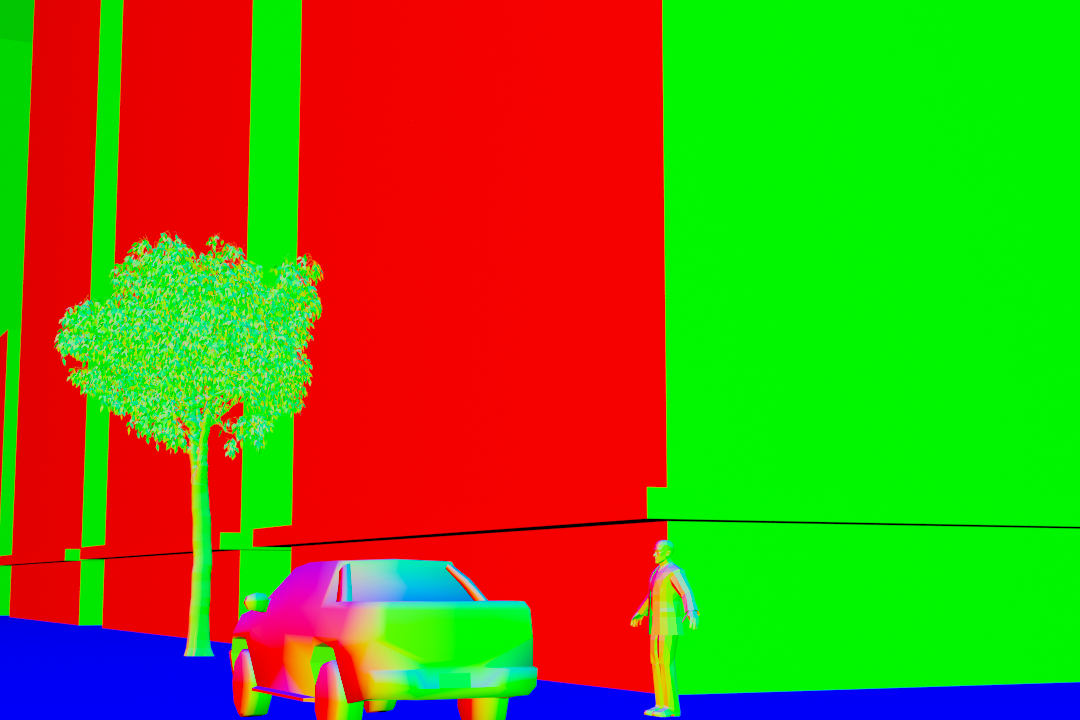}
\caption{Varying levels of fidelity and corresponding depth, surface normal, semantic annotations}
\label{fig_lof}
\end{figure}

\subsection{Rendering platform}
We developed a rendering platform, which exposes the parameters to scripting interface, to cop up with application domain modeling and sampling the domain models. 
The main goal of the graphics platform is solve rendering equation (to compute reflections from surface point), light transport equation (light computation through atmosphere), which are analytically intractable integrals.  The integrals are computed via random point-sample calculations that converge to the exact value of the integral with a variance that is reduced as $O(1/\theta_G)$ with the number of random samples. The error in the approximation of the integral appears in the final image as noise and in order to reduce the variance to acceptable levels the number of numerical samples ( $\theta_G$) used in this process has to be large.
Lambertian shading and Cook-Torrance shading (includes specular reflections for glassy surfaces) are simpler rendering methods and they do not model global interactions such as shadows, weather scattering etc. Please see Fig. \ref{fig_lof}.  In this work we simulate images with various degrees of model complexity to enable different levels of fidelity and analyze its impact on CNNs. 


\section{Experiments} \label{sec_exps}

As already mentioned, approximations made in domain models and rendering processes influence the statistics of rendered outputs, and, thus, biases the learned classifier. 
In this paper, we provide details of experiments to evaluate the bias in performance of the learned classifier on populations of input data from simulations with various levels-of-fidelity.
We postulate that the bias in performance would be dependent on local image statistics (e.g. the performance in boundaries between regions may be different compared to results from homogeneous regions) and use groundtruth information to analyze variability in bias due to local spatial contexts. Subsequently, we explore ways to reduce the performance bias through a combination of virtual world data and real-world data training.  We visualize histograms of image statistics in the training data against real data in order to get some insights on specific characteristics in input data statistics. 

\textbf{Baseline CNN architecture}:- Our baseline implementation is based on the DeepLab System \cite{chen2014semantic}.   DeepLab an modified version of VGG-16 net to operate at original image resolutions, by making following changes: (a) replace the fully connected layers with convolutional ones, (b) skip the last subsampling steps and upsample the feature-maps by applying hole algorithm. It still results coarser map with a stride of 8 pixels. Hence, targets during training are the ground truth labels subsampled by 8. During testing, bi-linear interpolation followed by fully connected conditional random field (CRF) was used to get final label maps. We modify the last layer of DeepLab from 21-class to 7-class (The classes include: building, pedestrian, trees, vehicles, ground, sky, and void). 

\textbf{Training}:- Although more simulated images could be used for full training, our models are initialized with pre-trained with PASCAL-context setup to skip longer training times. Stochastic gradient descent method and cross-entropy loss function are used with initial learning rate of 0.001, momentum of 0.9 and a weight decay of 0.0005. We use mini-batch of 4 images and learning rate is multiplied by 0.1 after every 2000 iterations. High-resolution input images are down-sampled by a factor 2. Training data is augmented by mirror reflections and random croppings from the original resolution images, which yields four times the data. We first train on training sets until the performance of validation set converges and then retrain on training+validation sets.  We use fixed parameters in the CRF inference process (10 mean field iterations with the default Gaussian potentials and Lagrangian weights provided by \cite{chen2014semantic}) in all reported experiments. 

\textbf{Comparison to Real world training}:-To evaluate the bias due to use of simulated data, we use a real world data set as a reference, which is \textit{CityScapes} \cite{cordts2016cityscapes}, recorded on the streets of several European cities. It provides a diverse set of videos with a public access to 3475 images (train-val) that has finer pixel-level annotations. We divide the database into two disjoint subsets for training, validation (3000 images, named as CS-train) and testing (475 image, named as CS-test) purposes. This dataset was adapted to the 7 class labels mentioned above. The CNN trained on this real data yields IoU 69.54\% on the test data. This is represented with a blue dotted line mark in Fig \ref{fig_gen_vs_realism}, which is a reference performance level for training-testing runs with simulated data.

\subsection{Level-of-Fidelity of Rendering}

We synthesized 5000 (4500 for training and 500 for validation purposes) samples from our domain model and rendered each image with different levels-of-fidelity. As progressive levels-of-fidelity, we use results from the: Lambertian shader (renders only diffuse reflections), Cook-Torrance shader (renders diffuse+glassy reflections), and Monte-Carlo path tracing (MCPT) (renders all global illumination such as shadows and weather effects such as scattering etc.) with increasing number of samples/pixel (numerical samples required to solve integral equations 
from 10 to 130 with a step-size of 30. We don't consider glassy reflections in MCPT as it apparently degrades the performance on CS-test. This means 7 sets of training data from same scene states, but with different levels-of-fidelity in rendering. Please see Fig \ref{fig_lof}, which are images corresponds to same scene state, but varying levels-of-fidelity. 

\begin{figure}
\centering
\begin{subfigure}[b]{0.327\textwidth}
\includegraphics[width=\textwidth,height=4cm]{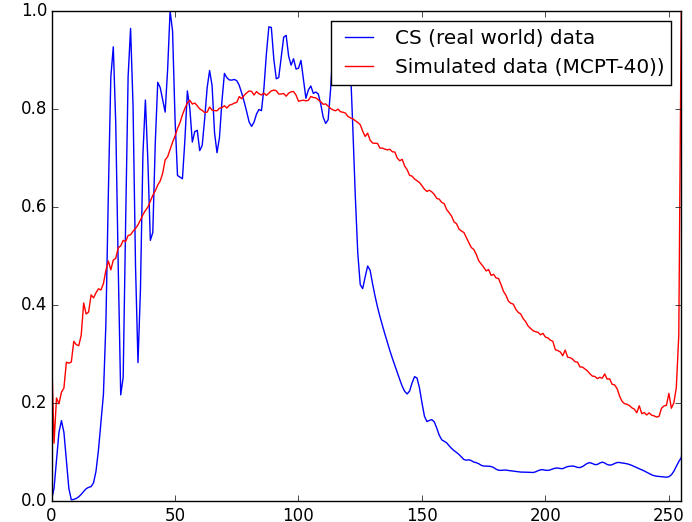}
\caption{\tiny{Deviations between intensity histograms of real and simulated datasets}}
\label{histograms_real_vs_virtual}
\end{subfigure}
\begin{subfigure}[b]{0.327\textwidth}
\includegraphics[width=\textwidth,height=4cm]{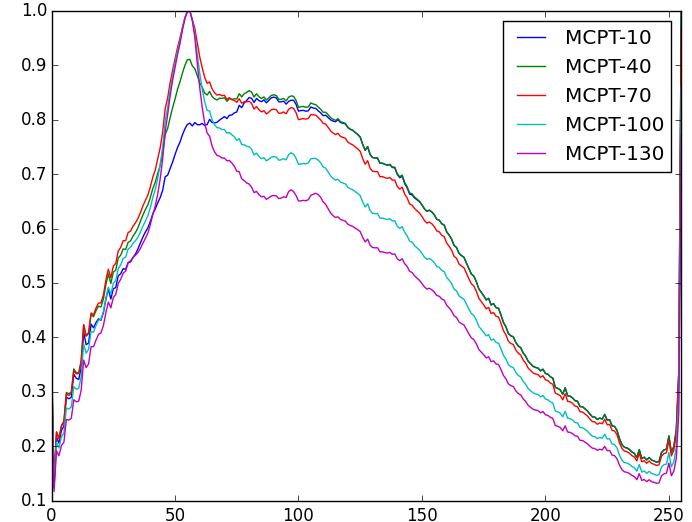}
\caption{\tiny{Deviations between histograms of simulated sets, due to level-of-fidelity}}
\label{histograms_sim_datasets}
\end{subfigure}
\begin{subfigure}[b]{0.327\textwidth}
\includegraphics[width=\textwidth,height=3.7cm]{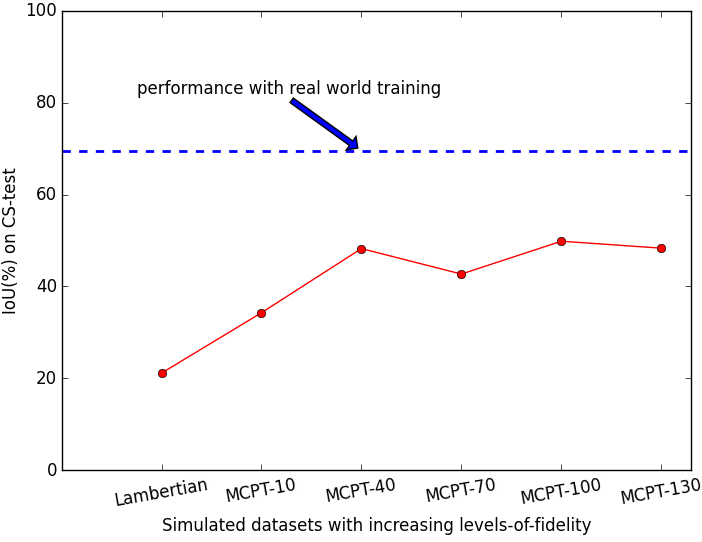}
\caption{\tiny{Impact of level-of-fidelity of simulations on the performance of CNN in real world. Blue dotted line marks the IoU for CNN trained with real world CS-train data}}
\label{fig_gen_vs_realism}
\end{subfigure}
\caption{Deviations propagate through rendered pixel space to CNN's output space}
\end{figure}

\textbf{Rendered data statistics}:- We start with comparing the statistics (histograms) of rendered data to know how the level-of-fidelity propagates to image intensity space. Please see Fig \ref{histograms_real_vs_virtual}, which displays the histograms (normalized) computed over all images of the sets. From the figure, in pixel intensity space, the intensity distributions of virtual and real worlds are quite varied. 

In Fig \ref{histograms_sim_datasets}, we illustrate histograms for gray levels normalized over large number of pixels of real data and simulated data. As expected, the modes 
of the histograms are sharper and pronounced as we use higher number of samples-per-pixel (note that the deviations between the histograms of MCPT-100 and MCPT-130 are quite less compared to other combinations). Hence, from Fig \ref{fig_lof} and Fig \ref{histograms_sim_datasets}, it is clear that there is variability due to the rendering pipeline sampling parameter. But, how do these deviations propagate through the training phase of a vision system like CNN ? 

\textbf{Bias in CNN performance}:- 
We train the CNN of DeepLab with each set and test its performance on the real world test data (CS-test). As shown in Fig \ref{fig_gen_vs_realism}, we plot the corresponding IoU \footnote{To assess performance, we use a standard performance metric, known as the intersection-over-union metric, $IoU = \frac{TP}{TP+FP+FN}$, where TP, FP, and FN are the numbers of true positive, false positive, and false negative pixels, respectively, determined over the whole test set.} measures to show how the generalization of the network (in real world) varies with level-of-fidelity in simulated training data. In all configurations, training with simulations seems worse compared to real world training. This bias could be due to level-of-details in the domain models and level-of-fidelity in the simulations. We only focus on level-of-fidelity for a moment. 

\textbf{Variations in CNN performance}:-
The IoU of CNN trained with simulated data using \textit{Lambertian} (diffuse reflections only) shader seems to be bad.  It is clear that diffuse material assumptions in simulations does not hold in most real world conditions.  When we used the \textit{Cook-Torrance} shader that simulates both diffuse and glassy reflections the performance was worse. An analysis of the error bars will need to be done to study the significance of this reduction. One possible explanation could be that CNN is not invariant to physical reflections. We therefore deactivated glassy reflections for this study and examined the effect of Monte-carlo sampling parameter in MCPT. 
These datasets are named as MCPT-$\theta_G$, where  $\theta_G$ being the number of numerical samples used per pixel. The plot seems to be more or less flat for MCPT datasets, especially after $\theta_G = 40$. The performance for MCPT varies between IoU $47 \pm 10$\%, mainly due to rendering noise in the simulations. From this experiment, we may conclude that CNN's seem to be less sensitive to level-of-fidelity in simulated training dataset used in our application context and that forty samples were enough for achieving good performance in our experiments. 
\begin{figure}
\centering
\begin{subfigure}[b]{0.38\textwidth}
\centering
\includegraphics[width=3.0cm,height=1.67cm]{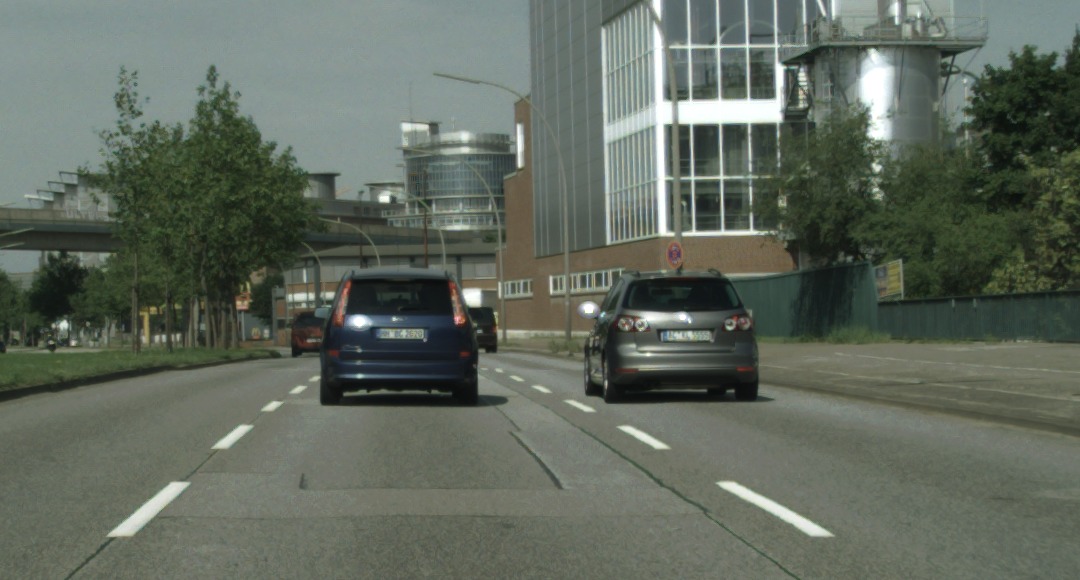}
\vfill
\includegraphics[width=3.0cm,height=1.67cm]{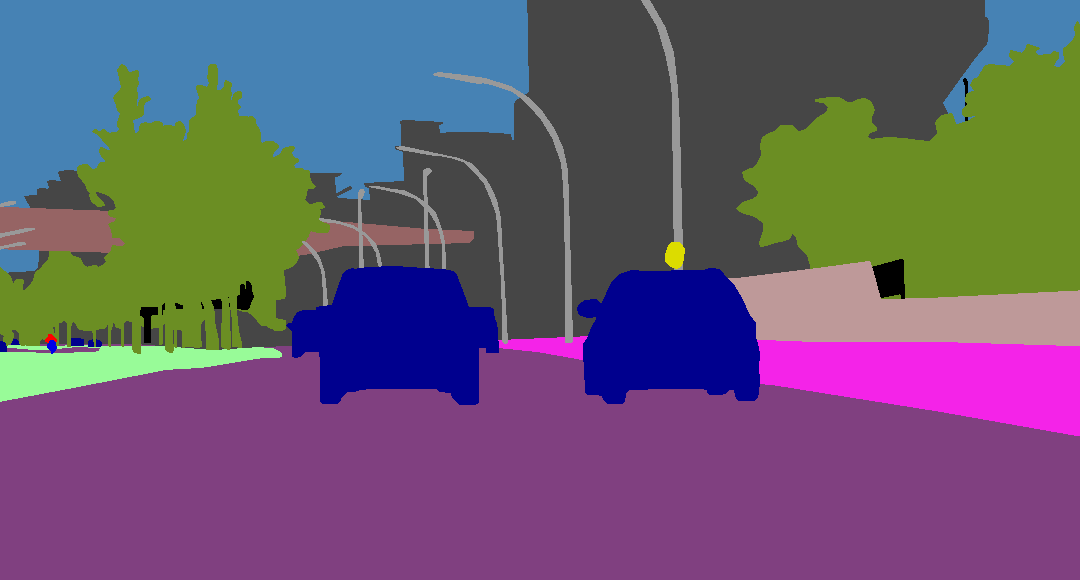}
\vfill
\includegraphics[width=3.0cm,height=1.67cm]{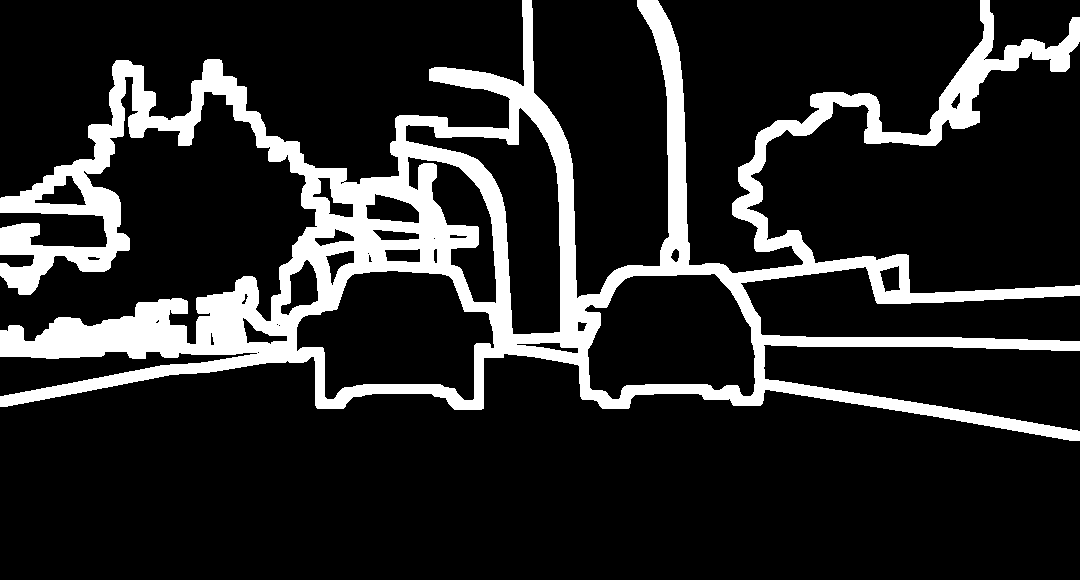}
\caption{\tiny{Top:Sample real world image, Middle: Semantic label-map, Bottom: Trimap of width 10 pixels}}
\label{trimap}
\end{subfigure}
\begin{subfigure}[b]{0.58\textwidth}
\includegraphics[width=\textwidth,height=5cm]{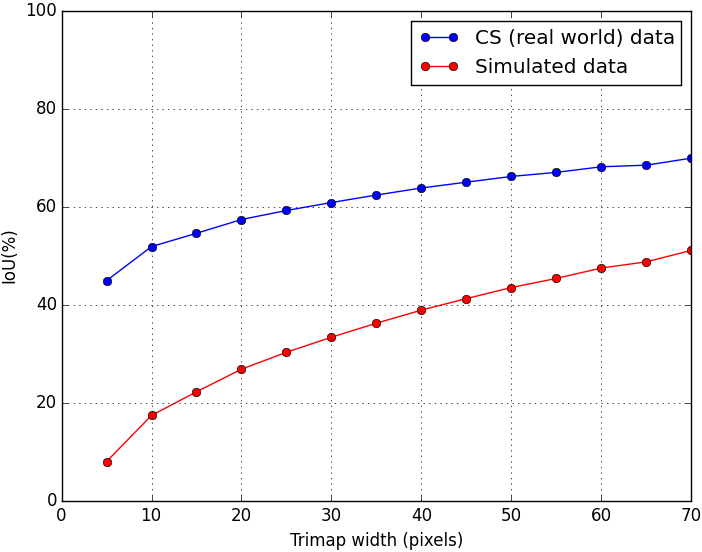}
\caption{\tiny{performance vs trimap width: performance between real and simulated worlds is more deviated near the boundaries} }
\label{performance_vs_trimap}
\end{subfigure}
\caption{Experiments with trimaps of varying widths conclude that the performance is more deviated near boundaries.}
\end{figure}
\subsection{Level-of-Detail in Domain models}
The performance of the trained system with only simulated data was always worse, compared to the setting of real world training (see the difference between red curve and blue line in the plot). \\
\textbf{Bias in Performance at Boundaries}:- To evaluate the variability in performance across spatial contexts in an image (e.g. near boundary pixels), we evaluate performance at object boundaries with the help of trimaps as explained in \cite{kohli2009robust}. Trimaps are masks with those pixels that are located within a narrow band of object boundaries, as shown in Fig \ref{trimap}. We create trimaps of varying pixel-widths for all test images, and compute IoU only at those pixels in the white band region. Fig \ref{performance_vs_trimap} show how performance (of real and virtual world training) changes from boundaries to more global spatial contexts. One can observe that the performance deviates more near boundaries (lower values of trimap's pixel-widths) than that over entire image space. We postulate that this may be due to the fact that the object boundaries and shadows in virtual worlds are quite sharp while real world boundaries have effects of color bleeding and penumbra (due to sensor effects). So, modeling sensor and lens effects in simulations may be important to reduce the statistical deviations in image space to pixel-level tasks such as semantic labeling. 

In this section, we consider the question: \textit{How can one reduce the performance bias by using a combination of simulated and real datasets?} 
\begin{wraptable}{r}{8cm}
\centering
\caption{\tiny{Comparison of generalization error for different training configurations}}\label{tab_real_vs_sim}
\tiny
\begin{tabular}{ccccccc}\\\toprule  
Architectue &Training &  Testing & IoU (\%) & IoU (\%) \\ 
  & data &  data & overall & building \\ \midrule
 CNN & CS-train &    CS-test & 69.54 & 80.43  \\ \hline
 CNN & Sim-5k &   CS-test & 42.09 & 60.02 \\ \hline
 CNN & Sim-20k &   CS-test & 46.38 &  65.75 \\ \hline
 CNN & Sim-20k+finetune &   CS-test & 68.13 & 76.63 \\ \hline
\end{tabular}
\end{wraptable}
Table \ref{tab_real_vs_sim} consolidates IoU measures on CS-test data for different training settings. The network achieves IoU of 69.54, when trained with CS-train. In the above experiments, the reported results are from the DCN trained with 5000 images simulated with MCPT-40. This model could achieve 42.09\% IoU. Although we increase train data size from 5000 to 20000 images, we could improve the performance hardly by 4\%.  The performance bias observed may be due to divergence in geometric and texture distributions between real world and our simulated worlds. We have used random 3D object shapes and textures, downloaded from online repositories. These models are low resolution polygonal mesh models with generic textures. But CityScapes is recorded in few European cities and European buildings have their own style of architecture and texture. Hence, this divergence in input domain models may reflect as bias in the network's performance. Thus we evaluated how the bias can be corrected with transfer learning concepts (fine-tuning of the classifier with some real world images) or learning domain models from real world data. 

\textbf{Domain adaptation can help}: If a few labelled real world samples are available, one can use them to fine tune the models that were trained with simulated data. We fine tune the above network (trained ion 20k simulated images) with just 10\% of CS-train data (300 images) and the result (denoted as sim-20k-finetune in the table) is on par with the one from training with full CS-train dataset. In conclusion, this allows to significantly save effort of large scale real world data collection and manual annotation while providing the systems of comparable performance with the systems, trained on a larger amount of manual annotations. In future work, we plan to bootstrap the domain models by preparing the texture-maps from the target real world data to reduce the divergence between virtual and real world distributions. If the real world samples are unlabelled, one can use co-training concepts to bootstrap the domain models and vision models simultaneously.

\section{Conclusions and Future directions} \label{sec_discuss}

In this work, we discussed the development of a probabilistic graphics simulation platform to tune/validate the computer vision systems. We validated our domain models and rendering processes in the platform with a specific example of urban scene semantic labeling with CNNs. Our experiments provide insights into the performance bias due to the level-of-fidelity of images as we vary the graphics rendering pipeline parameters. Experiments on simulated data indicate that CNN architectures trained were invariant to the level-of-fidelity after a minimum number of samples used in the monte-carlo integral computations in the renderer. In our specific case study, 40 samples/pixel were found to be good enough, although rendered images suffer from sampling noise. The performance bias seems to higher at boundaries and the magnitude of deviations decreases as we averaged over performance in the entire image space. Hence, virtual world training may be good for tasks like recognition (object detection and image classification) rather than fine pixel level segmentations. Our experiment with fine tuning concept shows that this performance bias can be corrected using combinations of simulated and real data sets, thus enabling reduction of the cost of data collection and manual annotation efforts.


\section*{Acknowledgements}

This work was supported by the German Federal Ministry of Education and Research (BMBF) in the projects, 01GQ0840 and 01GQ0841 (BFNT Frankfurt).

\small
\bibliographystyle{apalike}
\bibliography{egbib}

\end{document}